\documentclass{article}

\usepackage{microtype}
\usepackage{graphicx}
\usepackage{subcaption}
\usepackage{booktabs} 

\newcommand{\smallbrain}{%
    \ensuremath{\vcenter{\hbox{\includegraphics[height=2.5ex]{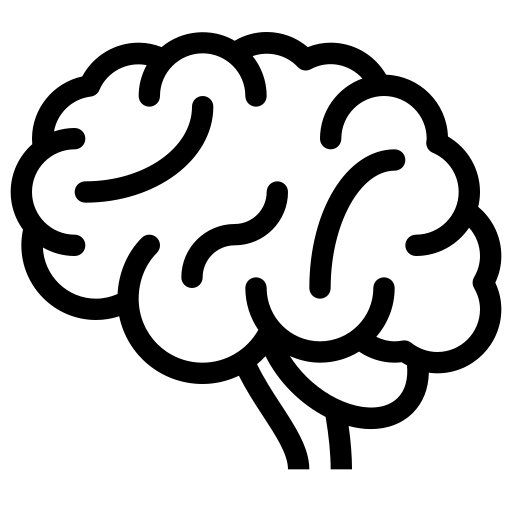}}}}%
}
\newcommand{\smallheart}{%
    \ensuremath{\vcenter{\hbox{\includegraphics[height=2.5ex]{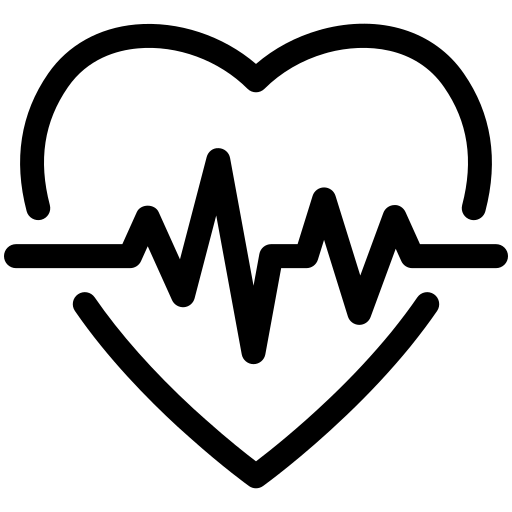}}}}%
}

\usepackage{hyperref}

\usepackage[preprint]{icml2026}

\usepackage{amsmath}
\usepackage{amssymb}
\usepackage{mathtools}
\usepackage{amsthm}

\usepackage[capitalize,noabbrev]{cleveref}

\theoremstyle{plain}

\theoremstyle{definition}

\theoremstyle{remark}

\usepackage[textsize=tiny]{todonotes}

\newcommand{\model}[0]{SleepFM$_{1}$}

\icmltitlerunning{}

\begin{document}

\twocolumn[
  \icmltitle{Pretraining on Sleep Data Improves non-Sleep Biosignal Tasks}



  \icmlsetsymbol{equal}{*}
  \begin{icmlauthorlist}
    \icmlauthor{William Lehn-Schiøler}{equal,BrainCapture,HealthTech,Compute}
    \icmlauthor{Magnus Ruud Kjær}{equal,HealthTech}
    \icmlauthor{Phillip Hempel}{equal,Got}
    \icmlauthor{Magnus Guldberg Pedersen}{BrainCapture,Compute}
    \icmlauthor{Rahul Thapa}{Stanford1}
    \icmlauthor{Bryan He}{Stanford2}
    \icmlauthor{Nicolai Spicher}{HealthTech}
    \icmlauthor{Andreas Brink-Kjaer}{HealthTech}
    \icmlauthor{Lars Kai Hansen}{Compute}
    \icmlauthor{Emmanuel Mignot}{Stanford}
  \end{icmlauthorlist}

  \icmlaffiliation{HealthTech}{DTU Health Tech, Technical University of Denmark, Kongens Lyngby, Denmark}
  \icmlaffiliation{Got}{University of Göttingen, Göttingen, Germany}
  \icmlaffiliation{Stanford1}
  {Department of Biomedical Data Science, Stanford University, Stanford, CA, USA}
  \icmlaffiliation{Stanford2}
  {Department of Computer Science, Stanford University, Stanford, CA, USA}
  \icmlaffiliation{Stanford}{Department of Psychiatry and Behavioral Sciences, Stanford University, Stanford, CA, USA}
  \icmlaffiliation{Compute}{DTU Compute, Technical University of Denmark, Kongens Lyngby, Denmark}
  \icmlaffiliation{BrainCapture}{BrainCapture ApS, Kongens Lyngby, Denmark}

  \icmlcorrespondingauthor{Emmanuel Mignot}{mignot@stanford.edu}

  \icmlkeywords{Machine Learning, ICML}

  \vskip 0.3in
]



\printAffiliationsAndNotice{}  


\begin{abstract}
Sleep foundation models have recently demonstrated strong performance on in-domain polysomnography tasks, including sleep staging, apnea detection, and disease risk prediction. In this work, we investigate whether sleep biosignals can serve as an effective pretraining distribution for learning representations that transfer beyond sleep to adjacent domains. Following sleep foundation models, we perform sleep-only multimodal contrastive pretraining (with a leave-one-out objective) and evaluate transfer to non-sleep EEG and ECG, two well-benchmarked biosignal modalities with heterogeneous datasets and clinically meaningful downstream tasks. Across eight downstream tasks spanning multiple EEG and ECG datasets, sleep pretraining consistently improves performance relative to training from scratch. Moreover, on several tasks, we achieve performance competitive with or surpassing prior specialized state-of-the-art and foundation models. 
\end{abstract}

\section{Introduction}


Sleep foundation models pretrained on large-scale polysomnography (PSG) data have recently been shown to perform well on sleep-specific tasks like sleep staging and sleep apnea severity detection \cite{thapa2024sleepfm,ruehland20112007,duce2014aasm} and some have gone beyond canonical tasks and focus on the prediction of future disease onset \cite{thapa2026multimodal, kjaer2025stanford}. PSG is the nocturnal recording of data containing coupled information across multiple biosignal domains, including brain activity (electroencephalography (EEG) and electrooculography (EOG)), respiration, cardiac activity (electrocardiography (ECG)), and muscle activity over long durations recorded synchronously. The PSG is the gold standard for diagnosing sleep apnea and determining sleep stages \cite{aasm_icsd}. 

In this study, we investigate the effect of sleep pretraining on other biosignal tasks during wakefulness. A priori, biosignals acquired during sleep might appear limited for this purpose because clinical PSG typically uses low-density EEG montages (e.g., a small set of frontal/central/occipital derivations) and only single-lead ECG due to acquisition constraints \cite{ruehland20112007,duce2014aasm}. Conversely, clinical EEG and ECG recordings often contain many more channels and leads for more fine-grained analysis; and they are recorded over short time-frames \cite{drew2004practice,klem1999ten}. We argue that the value of sleep data for representation learning lies less in spatial density and more in (1) long continuous contexts capturing various physiological states (e.g., sleep stages and transient events), (2) a broad representation of abnormal waveforms representing pathological processes (e.g., epileptic waveforms, cardiac arrhythmia, and shifts in relative signal frequency components) \cite{aasm_icsd}, and (3) tight cross-modal alignment between brain (EEG), heart (ECG), and breathing: Properties that provide rich self-supervised structure for learning transferable representations \cite{thapa2026multimodal}.

We build on the foundation model and pretraining approach introduced in \citet{thapa2026multimodal}.
Using this sleep-only pretraining, we evaluate transfer to two adjacent domains with distinct conventions and downstream tasks: wake EEG and ECG.

Both EEG and ECG are non-invasive medical tests that are part of a PSG but also find wide application as standalone modalities, e.g.\ for seizure detection (EEG) and arrhythmia detection (ECG).
EEG records the electrical activity of the brain using electrodes placed on the scalp. In recent years, EEG models have advanced rapidly, with latest approaches including LaBraM \cite{jianglarge}, CBraMod \cite{wangcbramod}, EEGMamba \cite{gui2024eegmamba}, and REVE \cite{elouahidi2025reve} --- all solving tasks on interictal events and seizure classification. ECG is a measurement of the cardiac electrical activity and is a cornerstone of cardiovascular medicine, finding wide application in primary and emergency care as well as prevention. ECG models typically rely on large ECG-only corpora and objectives tailored to detect abnormalities in cardiac morphology and rhythm, with supervised models achieving high accuracy \cite{narotamo2024deep, lunelli2025benchecgxecgbenchmarkbaseline}. 

\begin{figure*}[!t]
    \centering
    \includegraphics[width=\textwidth]{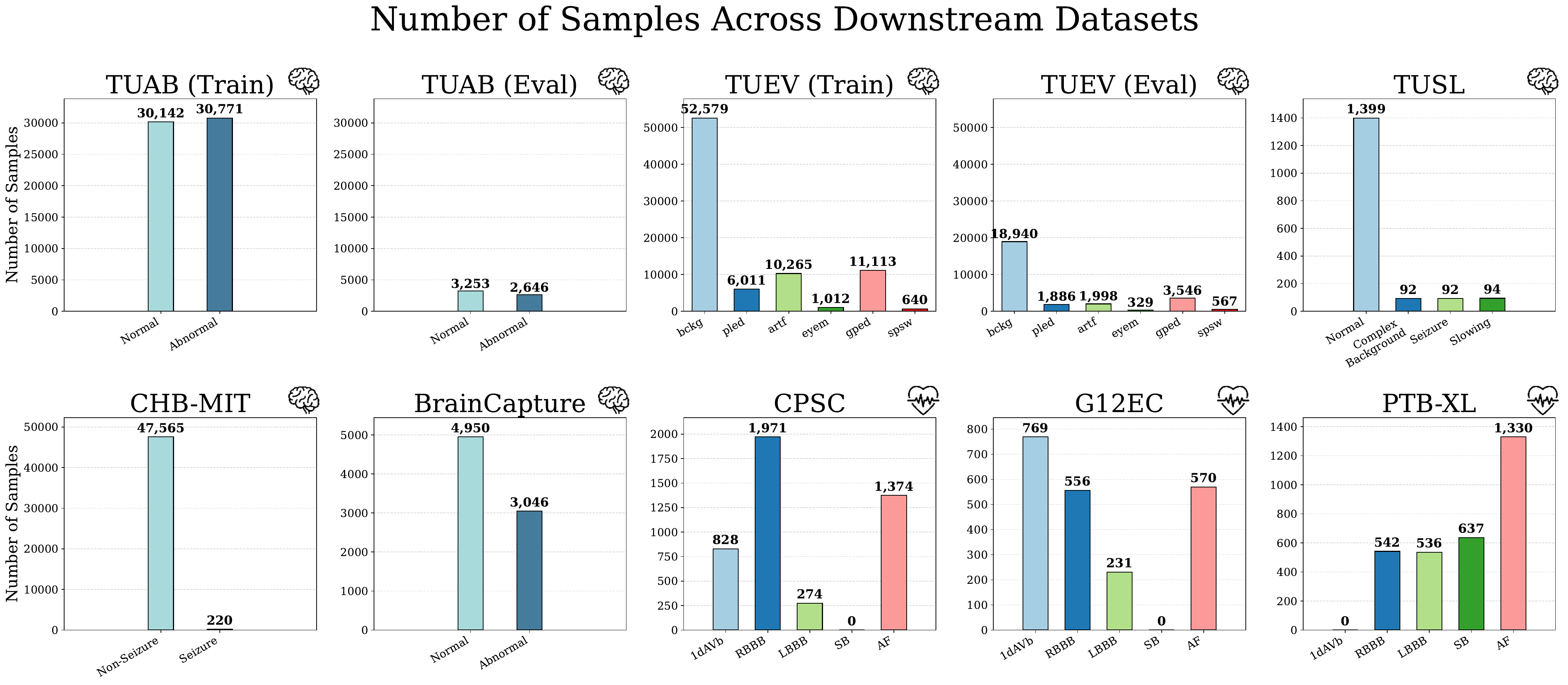}
    \caption{Overview of the study. \textbf{Left:} Sleep PSG is used for self-supervised pretraining of \model. \textbf{Right:} The pretrained encoder is fine-tuned on eight downstream tasks spanning five EEG and three ECG datasets. Class distributions for each downstream dataset are shown; EEG datasets are evaluated on 60-second windows, while ECG datasets use 10-second segments.}
    \label{fig:eeg_data_sample_dist}
\end{figure*}

\paragraph{Contributions.}
This paper asks a cross-domain question: Can sleep biosignals serve as an effective pretraining distribution for adjacent biosignal domains recorded during wakefulness? We evaluate transfer performance across eight downstream tasks spanning abnormality detection, seizure detection, interictal event classification, and arrhythmia classification on five EEG and three ECG datasets. We demonstrate that sleep pretraining consistently improves performance over non-pretrained baselines and achieves state-of-the-art results on multiple EEG benchmarks. Beyond the base SleepFM encoder \cite{thapa2026multimodal}, this work additionally contributes: (1) a channel positional encoding that injects 3D electrode geometry, enabling flexible channel-count inference; (2) evaluation on 8 downstream tasks (none evaluated in prior sleep foundation model work) across EEG and ECG; and (3) systematic channel ablation experiments demonstrating that 95\% performance is maintained with as few as 16--50\% of available channels.

\section{Related Work}

\subsection{Sleep foundation models and PSG pretraining}
PSG is an attractive biosignal pretraining source because it provides $>$8-hour-long, synchronized recordings of brain, cardiac, and respiratory physiology, enabling objectives that exploit both long temporal context and cross-modal alignment.  \citet{thapa2024sleepfm} introduced a sleep foundation model, demonstrating strong performance on canonical sleep tasks (sleep staging, apnea severity detection) and cross-modal retrieval, motivating sleep as a scalable representation-learning regime. 

Recent approaches developed a generative foundation model using score-based diffusion that enables zero-shot inference on arbitrary sensor combinations through Bayesian factorization of the score function, achieving near-human performance on sleep staging while maintaining flexibility across modalities \cite{van2024generative}. SynthSleepNet combined masked prediction with contrastive learning in a hybrid self-supervised framework, integrating a Mamba-based temporal context module to efficiently capture long-range dependencies and achieving strong performance on sleep staging, apnea, and hypopnea detection tasks, including in semi-supervised regimes with limited labels \cite{lee2025toward}. Stanford Sleep Bench \cite{kjaer2025stanford} provides a standardized PSG dataset and benchmark suite to systematically evaluate self-supervised pretraining methods for sleep foundation models, which was used to show that the cross-modal contrastive approach outperformed variations of masked and de-noising autoencoders across downstream tasks. 

Large-scale work further expanded the scope from sleep-centric tasks to broader clinical prediction and showed that sleep representations, when augmented with demographic metadata, can predict a wide range of future disease risks and mortality outcomes, highlighting PSG as an information-dense signal for general clinical outcome modeling \cite{thapa2026multimodal}. In the present work, no demographic features are used; all downstream gains are attributed strictly to biosignal representations.

\subsection{EEG foundation models}
EEG foundation modeling has advanced rapidly to address chronic label scarcity and the inherent heterogeneity across institutions, montages, and protocols. Early breakthroughs like BENDR \cite{Kostas2021BENDRUT} pioneered contrastive predictive coding for EEG, laying the groundwork for self-supervised learning in the domain. Building on this, LaBraM \cite{jianglarge} introduced a unified framework using neural tokenizers and masked signal modeling over channel patches to enable robust cross-dataset generalization, pretrained on the TUH EEG corpus (which includes TUAB and TUEV recordings, creating direct domain overlap with our downstream test sets). Similarly, BIOT \cite{yang2023biot} treats clinical EEG as a sequence of heterogeneous biosignals, utilizing a transformer-based tokenization that remains resilient to mismatched channels and variable sequence lengths.

Architectural specialization has further refined these models. CBraMod \cite{wangcbramod} employs a criss-cross transformer to disentangle spatial and temporal dependencies through masked reconstruction, while EEGMamba \cite{gui2024eegmamba} and FEMBA \cite{tegon2025fembaefficientscalableeeg} leverage State-Space Models (SSMs) to achieve efficient, long-sequence representation learning. LUNA \cite{doner2025luna} scales masked EEG pretraining to three model sizes (Base 7M, Large 43M, Huge 311M) and reports strong results on TUAB. Most recently, REVE \cite{elouahidi2025reve} introduced a 4D Fourier positional encoding scheme that combines 3D electrode coordinates with temporal patch indices, enabling pretraining on 60,000+ hours across 92 datasets with diverse electrode configurations.

While current efforts rely exclusively on in-domain (wake) EEG pretraining, our work investigates the transferability of \emph{sleep-only} pretraining. We demonstrate that despite the sparse EEG montages typical of PSG, the sheer scale and physiological richness of sleep data can effectively prime models for complex clinical EEG tasks.

\begin{table}[t]
\centering
\caption{Datasets used for downstream evaluation. EEG datasets are evaluated on 60-second windows, while ECG datasets use 10-second segments. TUAB and TUEV are pre-partitioned into training and evaluation directories. For ECG arrhythmia detection, Code-15 is used for fine-tuning, while PTB-XL, G12EC, and CPSC are used for external evaluation. Only one ECG per patient was included for each ECG dataset.}
\begin{tabular}{|l|c|c|c|}
\hline
\textbf{Dataset} & \textbf{Modality} & \textbf{\# Segments} & \textbf{\# Class} \\ \hline
TUAB & EEG & $60{,}913$ + $5{,}899$ & 2 \\ \hline
TUEV & EEG & $81{,}620$ + $27{,}266$ & 6 \\ \hline
TUSL & EEG & $1{,}633$ & 4 \\ \hline
CHB-MIT & EEG & $47{,}785$ & 2 \\ \hline
BrainCapture & EEG & $7{,}996$ & 2 \\ \hline \hline
Code-15 & ECG & $233{,}867$ & 6 \\ \hline
PTB-XL & ECG & $18{,}869$ & 4 \\ \hline
G12EC & ECG & $10{,}344$ & 4 \\ \hline
CPSC & ECG & $10{,}330$ & 4 \\ \hline
\end{tabular}
\label{tab:datasets}
\end{table}

\subsection{ECG SOTA models and supervised baselines}

ECG analysis has traditionally been dominated by supervised models trained for predefined clinical endpoints, such as rhythm or beat classification, using labels derived from expert annotation \cite{9190034}. Strong supervised baselines include convolutional architectures such as 1D-ResNets \cite{ribeiro2020automatic}, as well as recurrent models based on LSTMs and more recent variants such as xLSTM \cite{lunelli2025benchecgxecgbenchmarkbaseline}. More recently, state-space models \cite{mehari2022advancingstateoftheartecganalysis}, transformer-based architectures \cite{Yisimitila2026ECGTransformer}, and graph-based approaches \cite{Maurer2025xGNN4MI, 11339912} have further expanded the ECG modeling landscape. In parallel, several ECG foundation models have emerged through large-scale self-supervised pretraining \cite{almasud2025benchmarkingecgfoundationalmodels}.

However, many recent ECG results are reported under within-dataset evaluation protocols, challenge-style benchmarks, or task settings with different label spaces and preprocessing assumptions. In contrast, our ECG study is not intended as a comprehensive ECG benchmark, but as an external-validation transfer experiment: we fine-tune on Code-15 and evaluate on three external public datasets with aligned arrhythmia labels. We therefore use recent ECG literature primarily to contextualize our findings, while avoiding direct numerical comparisons where protocols are not sufficiently matched. We summarize recent ECG literature and ECG foundation models in Supplementary Table~\ref{tab:ecg_lit}.

\section{Methods}

We build on the representation learning framework introduced in \citet{thapa2026multimodal} to pretrain \model\ with two main differences from the original implementation: (1) we use patches of 1-second instead of 5-second patches, motivated by the need to preserve transient dynamics (e.g., sharp waves, R-peaks) that are important in non-sleep biosignal tasks but would be blurred at 5-second resolution; and (2) we use 2-minute windows for pretraining instead of 5-minute windows, which reduces memory requirements while still capturing multi-cycle physiological context. These design choices are intentionally minimal, allowing us to isolate the effect of the pretraining distribution itself rather than architectural specialization.
Subsequently, we fine-tune \model\ with a lightweight projector/classifier head for supervised downstream tasks.
To quantify the benefit of sleep-based pretraining for general biosignal modeling, we compare models initialized with pretrained encoder weights against identical architectures initialized with random weights.

\subsection{Datasets}

\paragraph{Pretraining datasets}

From 4 cohorts, Stanford Sleep Cohort (n = 24,137), Bioserenity (n = 18,869), MESA (n = 1,747), and MrOS (n = 3,340), \cite{thapa2026multimodal, hanif2024associations, zhang2018national, chen2015racial, blackwell2011associations}, we used 48,093 full night PSG recordings containing 432,000 hours of sleep data for pretraining using the training split from \cite{thapa2026multimodal}. All datasets were preprocessed following previous work \cite{thapa2026multimodal}. No demographic features were used during pretraining or fine-tuning; all reported gains are attributable to biosignal representations alone.

\paragraph{Downstream EEG datasets}
For EEG fine-tuning datasets, we choose both well-referenced and internal datasets with classification tasks including abnormality detection, event detection, slowing, and seizures; all datasets are described in \cref{tab:datasets} and \cref{fig:eeg_data_sample_dist}. Specifically, we fine-tune models on:
1) TUH EEG Abnormal (TUAB) \cite{lopez2015automated, obeid2016temple} with 2 classes: Normal and abnormal (2,563 subjects; 48.0 $\pm$ 17.9 years; 30\% female). The dataset is pre-partitioned into training and evaluation directories.
2) TUH EEG Events (TUEV) \cite{harati2015improved, obeid2016temple} with 6 classes: Spike and sharp wave (spsw), generalized periodic epileptiform discharges (gped), periodic lateralized epileptiform discharges (pled), eye movement (eyem), artifact (artf), and background (bckg).
3) TUH EEG Slowing (TUSL) \cite{shah2018temple, obeid2016temple} with 4 classes: Normal, complex background, slowing, and seizure (38 subjects).
4) The Children's Hospital Boston Dataset (CHB-MIT) \cite{PhysioNet-chbmit-1.0.0, shoeb2009application, goldberger2000physiobank} with 2 classes: Non-seizure and seizure (22 subjects; 9.9 $\pm$ 5.0 years; 78.3\% female; $<$1\% seizure prevalence).
5) BrainCapture \cite{armand2024low} with 2 classes: Normal and abnormal (2,340 subjects; 22.4 $\pm$ 18.9 years; 41\% female), recorded using a portable 27-channel device in resource-limited settings in Kenya.

Detailed dataset demographics and task descriptions are provided in \cref{tab:dataset_demographics} in the Appendix.

\vspace{3em}

\begin{figure*}[t]
    \centering
    \includegraphics[width=\textwidth]{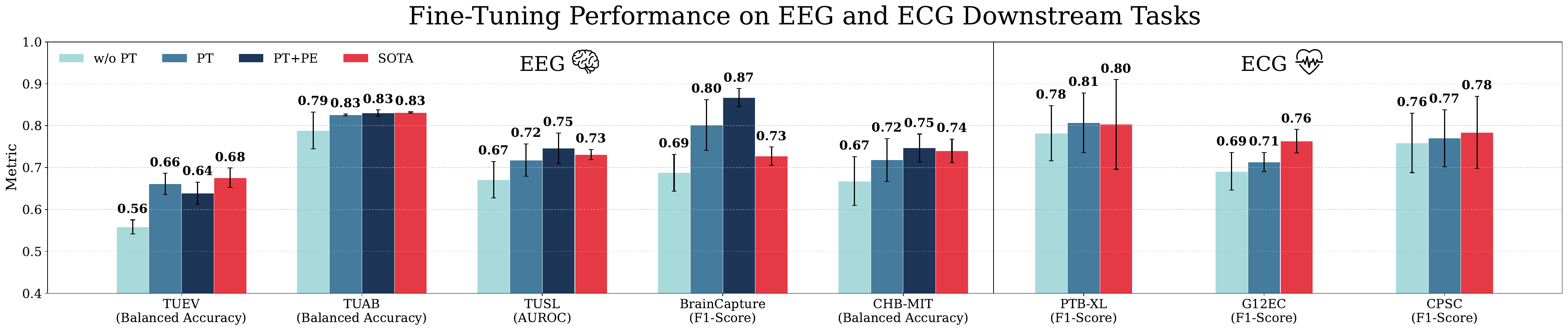}
    \caption{\model\ performance without pretraining (w/o PT) and with pretraining (PT) for ECG benchmarks, as well as with pretraining and positional encodings (PT + PE) across EEG benchmarks. Error bars denote 95\% confidence intervals: for EEG, intervals are computed across 15 independently trained models; for ECG, intervals are computed across the four arrhythmia detection tasks.}
    \label{fig:sleepfm-pt-vs-no-pt}
\end{figure*}

\paragraph{Downstream ECG datasets}
For ECG fine-tuning, the Code-15\% dataset was used, consisting of $345{,}779$ ECG measurements from $233{,}867$ patients acquired between 2010 and 2016 in the Telehealth Network of Minas Gerais, Brazil \cite{Lima2021ECGage}. Only the first available ECG per patient was used. For external evaluation, we used three well-benchmarked public ECG datasets comprising around $40{,}000$ ECG recordings from distinct patients across different healthcare systems and geographic regions (USA, Europe, and China): 1) PTB-XL \cite{9190034} ($18{,}869$ patients, Germany), 2) G12EC ($10{,}344$ patients, USA), and 3) CPSC \cite{PerezAlday2020PhysioNetChallenge} ($10{,}330$ patients, China). All ECGs were sampled at 500\,Hz.

For downstream evaluation, we formulated an arrhythmia detection task selecting the four most prevalent rhythm or conduction abnormalities per dataset. For PTB-XL: right bundle branch block (RBBB), left bundle branch block (LBBB), sinus bradycardia (SB), and atrial fibrillation (AF). For G12EC and CPSC: first-degree atrioventricular block (1dAVB), RBBB, LBBB, and AF.

\begin{table}[t]
\centering
\caption{The five components of the model used to learn the downstream EEG tasks. \textbf{The complete model has 3.8M parameters.}}
\begin{tabular}{|l|l|}
\hline
\textbf{Component}      & \textbf{\# Parameters} \\ \hline
\model\ Encoder  & 3.0M \\ \hline
Positional Encoder  & 100K \\ \hline
Masked Contextualizer & 101K\\ \hline
Temporal Aggregator     & 0 \\ \hline
Non-Linear Projection        & 592K \\ \hline
Classification Head     & 4.6K\\ \hline
\end{tabular}
\label{tab:eeg_model_components}
\end{table}

\subsection{Pretraining}

\model\ uses the architecture from \citet{thapa2026multimodal}, which produces embeddings of dimension 128. We specify the signal patching length to 1-second and input 2-minute windows.
The encoder supports two output granularities: (i) a single embedding per input window, yielding a compact representation of shape $(1,128)$, or (ii) a sequence of embeddings at 1-second resolution, yielding shape $(120,128)$ for a 2-minute window.
In all experiments, we use the 1-second resolution embeddings, as they retain temporally localized information required for short-event biosignal classification.

We intentionally make only minimal adaptations to the SleepFM encoder to ensure compatibility with short-duration, non-sleep biosignal events. These simple design changes largely isolate the effect of the pretraining distribution itself, allowing us to attribute downstream gains to sleep-based representation learning rather than architectural specialization.

\begin{table}[t]
\centering
\caption{The components of the model used to learn the downstream ECG tasks. \textbf{The complete model has 3.6M parameters.}}
\begin{tabular}{|l|l|}
\hline
\textbf{Component}      & \textbf{\# Parameters} \\ \hline
\model\ Encoder  & 3.0M \\ \hline
Classification Head  & 575.5K \\ \hline
\end{tabular}
\label{tab:ecg_model_components}
\end{table}

\begin{table}[t]
\centering
\caption{Key hyperparameters used for EEG and ECG fine-tuning.}
\label{tab:hyperparams}
\small
\begin{tabular}{|l|c|c|}
\hline
\textbf{Hyperparameter} & \textbf{EEG} & \textbf{ECG} \\ \hline
Input window & 60 s & 30.72 s \\ \hline
Sampling rate & 128 Hz & 125 Hz \\ \hline
Bandpass filter & 1--70 Hz & None \\ \hline
Optimizer & AdamW & AdamW \\ \hline
Learning rate & 1e-4 & 1e-4 \\ \hline
Batch size & 32 & 64 \\ \hline
Fine-tuning strategy & Full & Full \\ \hline
Validation & 15-fold CV & External datas \\ \hline
\end{tabular}
\end{table}

\begin{table*}[ht]
\centering
\caption{Results on TUH EEG Events \smallbrain\ (TUEV, 6 classes). We benchmark \model\ without pretraining (PT), with pretraining, with pretraining and positional encoding (PE), and with a linear probe (frozen encoder). Note: LaBraM was pretrained on the TUH EEG corpus, which includes TUEV recordings, creating direct domain overlap with the test set. Our model has no such overlap.}
\label{tab:tuev_results}
\begin{tabular}{lcccc}
\hline
\textbf{Model} & \textbf{Model Size} & \textbf{Balanced Accuracy} & \textbf{Weighted F1} & \textbf{Cohen's $\kappa$} \\
\hline
BIOT  \cite{yang2023biot}         & 3.2M   & $0.528\pm0.023$ & $0.749\pm0.008$ & $0.527\pm0.025$ \\
LaBraM-Base  \cite{jianglarge}    & 5.8M   & $0.641\pm0.007$  & $0.831\pm0.005$ & $0.664\pm0.009$ \\
LaBraM-Large \cite{jianglarge}    & 46M    & $0.658\pm0.016$  & $0.832\pm0.004$ & $0.662\pm0.014$ \\
LaBraM-Huge  \cite{jianglarge}    & 369M   & $0.662\pm0.017$  & $0.833\pm0.009$ & $0.675\pm0.020$ \\
CBraMod       \cite{wangcbramod}  & 5.8M   & $0.667\pm0.011$  & $0.834\pm0.006$ & -- \\
REVE-Base    \cite{elouahidi2025reve} & 69.2M & $\mathbf{0.676\pm0.023}$ & $\mathbf{0.845\pm0.013}$ & -- \\
\hline
\model\ w/o PT                    & 3.7M   & $0.559\pm0.017$ & $0.738\pm0.016$ & $0.497\pm0.046$ \\
\model\ w. PT                     & 3.7M   & $0.661\pm0.025$ & $0.821\pm0.011$ & $0.520\pm0.068$ \\
\model\ w. PT + PE                & 3.8M   & $0.639\pm0.027$ & $0.808\pm0.016$ & $0.604\pm0.050$ \\
\model\ Linear Probe              & 3.7M   & $0.485\pm0.031$ & --              & $0.441\pm0.024$ \\
\hline
\end{tabular}
\end{table*}

\begin{table*}[t]
\centering
\caption{Performance on TUH EEG Slowing \smallbrain\ (TUSL, 4 classes). REVE, LaBraM, CBraMod, and BIOT do not report results on TUSL in their original publications, and their public checkpoints are not configured for this dataset; we therefore omit them from this comparison. We benchmark \model\ without pretraining (PT), with pretraining, and with pretraining and positional encoding (PE).}
\label{tab:tusl_results}
\begin{tabular}{lccccc}
\hline
\textbf{Model} & \textbf{Model Size} & \textbf{AUROC} & \textbf{AUC-PR} \\
\hline
BrainBERT  \cite{wang2023brainbertselfsupervisedrepresentationlearning} & 43.2M  & $0.588\pm0.013$ & $0.352\pm0.003$ \\
EEGFormer-Base   \cite{chen2024eegformertransferableinterpretablelargescale} & 2.3M   & $0.713\pm0.010$ & $0.393\pm0.003$  \\
FEMBA-Base   \cite{tegon2025fembaefficientscalableeeg} & 47.7M  & $0.731\pm0.012$ & $0.289\pm0.009$ \\
\hline
\model\ w/o PT                        & 3.7M     & $0.671\pm0.044$ & $0.381\pm0.026$  \\
\model\ w. PT                         & 3.7M     & $0.718\pm0.039$ & $0.421\pm0.028$  \\
\model\ w. PT + PE                    & 3.8M     & $\mathbf{0.746\pm0.037}$ & $\mathbf{0.468\pm0.030}$  \\
\hline
\end{tabular}
\end{table*}

\begin{table*}[t]
\centering
\caption{Results on BrainCapture \smallbrain\ (2 classes). We benchmark \model\ without pretraining (PT), with pretraining, with pretraining and positional encoding (PE), and with a linear probe (frozen encoder).}
\label{tab:braincapture_results}
\begin{tabular}{lcccc}
\hline
\textbf{Model} & \textbf{Model Size} & \textbf{Balanced Accuracy} & \textbf{Weighted F1} & \textbf{Cohen's $\kappa$}  \\
\hline
BENDR  \cite{Kostas2021BENDRUT}            & 70.9M  & $0.637\pm0.035$ & $0.727\pm0.022$ & -- \\
LaBraM    \cite{jianglarge}                & 5.8M   & $0.745\pm0.026$ & $0.754\pm0.084$ & -- \\
\hline
\model\ w/o PT      & 3.7M & $0.762\pm0.040$ & $0.688\pm0.044$ & -- \\
\model\ w. PT       & 3.7M & $0.857\pm0.056$ & $0.802\pm0.060$ & -- \\
\model\ w. PT + PE  & 3.8M & $\mathbf{0.915\pm0.007}$ & $\mathbf{0.867\pm0.022}$ & -- \\
\model\ Linear Probe & 3.7M & $0.805\pm0.027$ & -- & $0.578\pm0.065$ \\
\hline
\end{tabular}
\end{table*}

\begin{table*}[t]
\centering
\caption{Results on TUH EEG Abnormal \smallbrain\ (TUAB, 2 classes). We benchmark \model\ without pretraining (PT), with pretraining, with pretraining and positional encoding (PE), and with a linear probe (frozen encoder). Note: LaBraM and LUNA were pretrained on wake EEG data with direct domain overlap with TUAB; our model was pretrained exclusively on sleep PSG with no such overlap.}
\label{tab:tuab_results}
\begin{tabular}{lcccc}
\hline
\textbf{Model} & \textbf{Model Size} & \textbf{Balanced Accuracy} & \textbf{AUC-PR} & \textbf{AUROC} \\
\hline
BIOT \cite{yang2023biot}           & 3.2M   & $0.796\pm0.006$ & $0.879\pm0.002$ & $0.882\pm0.004$ \\
LUNA-Base \cite{doner2025luna}     & 7M     & $0.806\pm0.001$ & -- & -- \\
LUNA-Large \cite{doner2025luna}    & 43M    & $0.810\pm0.001$ & -- & -- \\
LaBraM-Base \cite{jianglarge}      & 5.8M   & $0.814\pm0.002$ & $0.897\pm0.002$ & $0.902\pm0.001$ \\
LUNA-Huge \cite{doner2025luna}     & 311.4M & $0.816\pm0.001$ & -- & -- \\
LUNA-Base \cite{doner2025luna}     & 7M     & $0.806\pm0.080$ & $0.895\pm0.002$ & $0.887\pm0.002$ \\
LaBraM-Large \cite{jianglarge}     & 46M    & $0.823\pm0.002$ & $0.913\pm0.001$ & $0.913\pm0.001$ \\
REVE-Base \cite{elouahidi2025reve} & 69.2M  & $\mathbf{0.832\pm0.001}$ & $\mathbf{0.928\pm0.001}$ & $\mathbf{0.925\pm0.001}$ \\
LaBraM-Huge \cite{jianglarge}      & 369M   & $0.826\pm0.001$ & $0.920\pm0.001$ & $0.916\pm0.002$ \\
\hline
\model\ w/o PT                     & 3.7M   & $0.788\pm0.055$ & $0.862\pm0.005$ & $0.860\pm0.006$ \\
\model\ w. PT                      & 3.7M   & $0.807\pm0.039$ & $0.898\pm0.002$ & $0.891\pm0.002$ \\
\model\ w. PT + PE                 & 3.8M   & $0.825\pm0.007$ & $0.925\pm0.010$ & $0.910\pm0.006$ \\
\model\ Linear Probe               & 3.7M   & $0.796\pm0.006$ & -- & -- \\
\hline
\end{tabular}
\end{table*}

\begin{table*}[t]
\centering
\caption{Results on CHB-MIT seizure detection \smallbrain\ (2 classes). We benchmark \model\ without pretraining (PT), with pretraining, with pretraining and positional encoding (PE), and with a linear probe (frozen encoder). Note: Cohen's $\kappa \approx 0$ for all strategies reflects extreme class imbalance ($<$1\% seizure prevalence), not a failure of representation quality.}
\label{tab:chbmit_results}
\begin{tabular}{lcccc}
\hline
\textbf{Model} & \textbf{Model Size} & \textbf{Balanced Accuracy} & \textbf{AUC-PR} & \textbf{AUROC} \\
\hline
BIOT    \cite{yang2023biot}         & 3.2M   & $0.707\pm0.046$ & $0.328\pm0.046$ & $0.876\pm0.028$ \\
LaBraM-Base   \cite{jianglarge}     & 5.8M   & $0.706\pm0.036$ & $0.329\pm0.040$ & $0.868\pm0.020$ \\
CBraMod \cite{wangcbramod}          & 4.0M   & $0.740\pm0.028$ & $0.369\pm0.038$ & $\textbf{0.889}\pm\textbf{0.015}$ \\
\hline
\model\ w/o PT  & 3.7M & $0.668\pm0.058$ & $0.203\pm0.070$ & $0.829\pm0.102$ \\
\model\ w. PT   & 3.7M & $0.718\pm0.051$ & $0.436\pm0.080$ & $0.836\pm0.052$ \\
\model\ w. PT + PE & 3.8M & $\textbf{0.747}\pm\textbf{0.033}$ & $\textbf{0.456}\pm\textbf{0.052}$ & $0.885\pm0.037$ \\
\model\ Linear Probe & 3.7M & $0.682\pm0.121$ & -- & -- \\
\hline
\end{tabular}
\end{table*}

\subsection{Finetuning}

We fine-tune the \model\ encoder together with a lightweight projector/classifier head for supervised downstream tasks. All encoder parameters are updated during fine-tuning (full fine-tuning). To quantify the benefit of sleep-based pretraining, we compare models initialized with pretrained encoder weights against identical architectures initialized with random weights. As an additional baseline, we also report linear probe results (frozen encoder, trainable head only) to assess the quality of the pretrained representations independently of fine-tuning.

Baseline models (LaBraM, BIOT, CBraMod, etc.) are compared using numbers reported in their original publications, evaluated on the same fixed test sets. Fine-tuning details are summarized in Table~\ref{tab:hyperparams}.

\paragraph{EEG Downstream Tasks}

For all downstream EEG tasks, we fine-tune \model\ on 60-second windows, using per-second encodings of size $(60, 128)$.

The complete model used for downstream EEG tasks has 3.8M parameters and is composed of five components (Table~\ref{tab:eeg_model_components}): the \model\ transformer encoder, a masked contextualizer (a lightweight conditioning block applying masking, positional convolution, and 1$\times$1 projection to prepare encoder embeddings for pooling, following BENDR \cite{Kostas2021BENDRUT}), a temporal aggregator (mean pooling; no learnable parameters), a non-linear projection, and a classification head.

Preprocessing followed an evolved version of the SPEED framework \cite{speed_preprocessing}: resampling to 128\,Hz, 1--70\,Hz bandpass filter, and line noise filter at 50 or 60\,Hz. For annotations shorter than 60 seconds, we extend the window beyond the end of the annotation.

To address class imbalance, we employ targeted upsampling of minority classes to parity with the majority class, combined with stochastic spatial channel dropout during upsampled epochs to prevent overfitting and promote channel-invariant representations.

To maintain methodological consistency with prior work, we adapt validation strategy to dataset structure. For TUAB and TUEV, which are pre-partitioned into training and evaluation directories, we perform 15 independent experimental runs with unique random seeds. For remaining datasets without fixed partitions, we use 15-fold cross-validation.

\paragraph{Positional Encoding}

During fine-tuning on downstream EEG tasks, we explore REVE-inspired channel positional encodings (denoted PE in figures and tables) using only the spatial component (x, y, z coordinates) without temporal encoding. We augment the montage-agnostic embeddings from \model's convolutional tokenizer with 3D Fourier positional encodings based on standard 10-20 system electrode locations, adding this spatial information after tokenization but before spatial pooling. For bipolar channels, the channel location is computed as the average of the constituent electrode positions.

\paragraph{ECG Downstream Tasks}

For ECG downstream tasks, we fine-tune the pretrained \model\ encoder with a lightweight classification head for multi-label prediction. The complete model has 3.6M parameters (Table~\ref{tab:ecg_model_components}).

All ECG recordings were resampled to 125\,Hz in a fixed lead order across datasets. Inputs were converted to fixed-length segments of 3,840 samples (30.72\,s) via symmetric zero-padding tracked with a binary mask. The encoder tokenizer splits each segment into 30 non-overlapping tokens of 128 samples ($\approx$1.02\,s per token). No additional signal preprocessing (filtering, baseline-wander removal, amplitude standardization) was applied.

\subsection{Channel ablation}
To evaluate robustness of our channel-agnostic pretraining and fine-tuning approach, we assess model performance when only a subset of channels is available at test time. For EEG datasets, we randomly sample channel subsets ranging from 1 to 19 channels, performing 5 independent draws per subset size. For ECG datasets, we evaluate across systematically selected lead subsets ranging from 1 to 12 leads.

\section{Results}

\begin{table*}[t]
\centering
\caption{F1 scores across PTB-XL, G12EC, and CPSC \smallheart\ datasets. \model\ is fine-tuned on Code-15 and evaluated on these external datasets; this is an external-validation transfer experiment, not an in-distribution benchmark. Bolded values indicate the best result per column.}
\label{tab:ecg_f1_combined_results}
\begin{tabular}{lccc}
\hline
\textbf{Model} & \textbf{PTB-XL} & \textbf{G12EC} & \textbf{CPSC} \\
\hline
Resnet1d\_wang \cite{7966039} & $0.723\pm0.233$ & $0.728\pm0.070$ & $0.748\pm0.113$ \\
Xresnet1d101 \cite{8954382} & $0.803\pm0.107$ & $\textbf{0.763}\pm\textbf{0.028}$ & $0.758\pm0.101$ \\
Inception Time \cite{ismail_fawaz_inceptiontime_2020} & $0.771\pm0.089$ & $0.736\pm0.056$ & $0.770\pm0.070$ \\
1DResNet \cite{ribeiro2020automatic} & $0.775\pm0.100$ & $0.737\pm0.041$ & $\textbf{0.784}\pm\textbf{0.086}$ \\
\hline
\model\ w/o PT & $0.782\pm0.066$ & $0.691\pm0.045$ & $0.759\pm0.071$ \\
\model\ w. PT & $\textbf{0.807}\pm\textbf{0.071}$ & $0.713\pm0.023$ & $0.770\pm0.068$ \\
\hline
\end{tabular}
\end{table*}

\begin{figure*}[t]
    \centering
    \includegraphics[width=2\columnwidth]{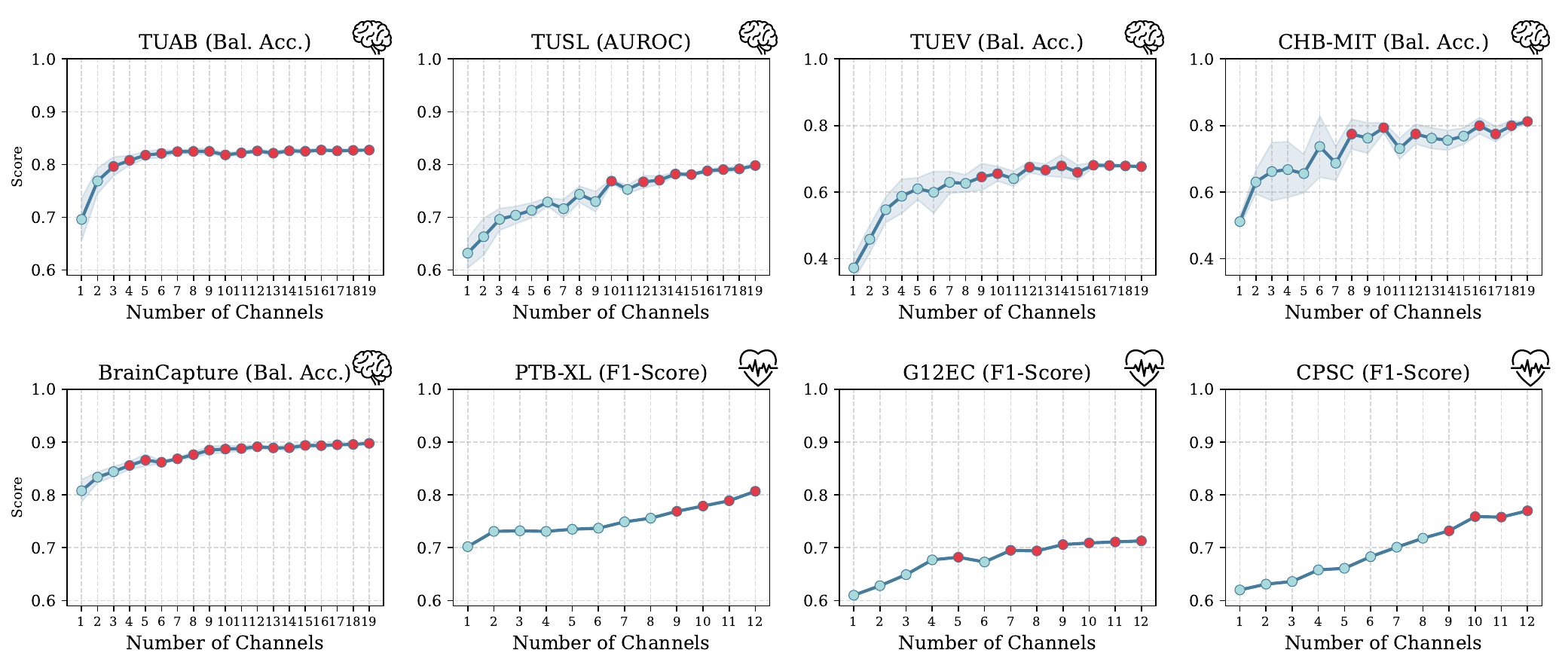}
    \caption{\model\ performance with pretraining scaled by the number of input channels on the EEG and ECG datasets. Red points denote when the performance is at or has surpassed 95\% of the performance on all channels.}
    \label{fig:EEG_scaling}
\end{figure*}

Across all eight downstream tasks spanning five EEG and three ECG (\cref{tab:datasets}) datasets, sleep-only pretraining consistently improves performance relative to training from scratch (\cref{fig:sleepfm-pt-vs-no-pt}). These gains hold across heterogeneous task types --- including abnormality detection, event classification, seizure detection, and arrhythmia classification --- and across datasets with different channel configurations and recording protocols. Notably, improvements are larger in EEG tasks, where labeled data are scarcer and datasets are more heterogeneous, suggesting that sleep pretraining is particularly effective in low- to moderate-data regimes.

The linear probe results (Tables~\ref{tab:tuev_results}--\ref{tab:chbmit_results}) show that the sleep-pretrained encoder already produces informative representations without any fine-tuning: on binary tasks, linear probes achieve TUAB balanced accuracy of 0.796 and BrainCapture balanced accuracy of 0.805. Full fine-tuning provides meaningful additional gains (e.g., +0.029 balanced accuracy on TUAB, +0.126 on BrainCapture), consistent with standard transfer learning findings. The near-zero Cohen's $\kappa$ for CHB-MIT across all strategies reflects the extreme class imbalance ($<$1\% seizure prevalence) rather than a failure of representation quality.

\subsection{Sleep pretraining is competitive with domain-specific state of the art}

Despite being pretrained exclusively on sleep PSG data, \model\ achieves performance that is competitive with --- and in several cases surpasses --- recent EEG foundation models pretrained directly on in-domain wake EEG data. On TUAB, our 3.8M-parameter model with PT+PE (balanced accuracy 0.825) outperforms all three LUNA model sizes including LUNA-Huge (311M parameters, BA 0.816) and LaBraM-Large (46M, BA 0.823), and is within rounding margin of LaBraM-Huge (369M, BA 0.826). This is particularly notable because LaBraM and LUNA were pretrained on the TUH EEG corpus, which includes TUAB recordings --- giving them direct domain overlap with the test set that our model does not share. On TUEV, our model (BA 0.639) is essentially tied with LaBraM-Base (BA 0.641) despite the same domain-overlap disadvantage. \model\ achieves state-of-the-art results on TUSL and CHB-MIT, outperforming FEMBA-Base, EEGFormer, LaBraM, and BIOT on those benchmarks (Tables~\ref{tab:tuev_results}--\ref{tab:chbmit_results}). On BrainCapture, \model\ with PT+PE substantially outperforms both BENDR and LaBraM.

In ECG tasks, our evaluation is an external-validation transfer experiment (fine-tuned on Code-15, evaluated on PTB-XL, G12EC, CPSC with aligned arrhythmia labels) rather than a comprehensive ECG benchmark. Under this stricter generalization test, sleep pretraining consistently improves over no pretraining and is competitive with supervised baselines trained on large ECG-only corpora, outperforming Inception Time and 1DResNet on PTB-XL (\cref{tab:ecg_f1_combined_results}). ECG gains are smaller than EEG gains; we discuss this in Section~\ref{sec:discussion}.

\subsection{Channel-agnostic pretraining yields high performance on channel subsets}

Beyond aggregate performance, we evaluate robustness to channel availability. \model\ maintains high performance even when a substantial fraction of EEG channels are removed at test time. Across EEG tasks, 16--50\% of channels are required to reach 95\% of the full-channel performance (\cref{fig:EEG_scaling}). Similarly, on ECG tasks, \model\ requires only 41--75\% of leads to achieve 95\% performance. This channel efficiency emerges without any channel-specific pretraining or architectural assumptions, highlighting the flexibility of sleep-pretrained representations and suggesting a path toward less burdensome, lower-cost recording systems.

\section{Discussion}
\label{sec:discussion}

Our results demonstrate that \model\ matches or outperforms established EEG foundation models like BIOT, LaBraM, and CBraMod, establishing sleep-only PSG as a pretraining corpus with strong transfer potential. While \model\ does not reach the absolute accuracy of the state-of-the-art REVE, it is important to note that REVE utilizes a 20-fold increase in parameter count. When \model\ is augmented with REVE-style channel-level positional encoding during fine-tuning, the performance gap narrows substantially, suggesting that the physiological richness of longitudinal sleep recordings provides a highly transferable feature set for clinical EEG tasks.

\paragraph{Why does sleep pretraining transfer?}
We offer several mechanistic observations. First, sleep EEG spans a wider frequency range than most wake EEG paradigms, including the delta--theta activity common to pathological EEG (e.g., slowing in encephalopathy, seizure ictal patterns). Second, long-context pretraining windows (2 minutes) capture temporal dependencies at the scale of clinical events such as seizures and burst suppression, which shorter-window models miss. Third, the multimodal contrastive objective over EEG, EOG, EMG, and SpO$_2$ forces the encoder to disentangle EEG-specific features from peripheral physiology, yielding representations that are more specific to neural dynamics. Together, these properties explain why a model trained without any wake EEG can produce representations competitive with --- or superior to --- models pretrained directly on the target domain.

\paragraph{ECG transfer.}
We emphasize that the ECG study is designed as an external-validation transfer experiment, not as a comprehensive ECG benchmark: models are fine-tuned on Code-15 and evaluated on three external public datasets with aligned arrhythmia labels. We therefore interpret the ECG findings as evidence of positive transfer under distribution shift, rather than as a definitive state-of-the-art comparison against the full contemporary ECG literature (see Appendix Table~\ref{tab:ecg_lit} for context).

ECG gains were smaller than EEG gains. This likely reflects both stronger domain shift in the ECG evaluation and the fact that sleep recordings are more directly centered on EEG phenomena. Concretely, PSG records single-lead ECG, so the representation learning signal for cardiac morphology is sparser and potentially noisier than for EEG. This limits, but does not eliminate, transfer: sleep pretraining still consistently improves over random initialization across all ECG datasets.

\paragraph{Channel robustness.}
The resilience of \model\ to significant channel pruning indicates a robust capacity for capturing inter-channel correlations. Across EEG tasks, as few as 16--50\% of electrodes are sufficient to retain 95\% of full-channel performance; on TUAB and BrainCapture specifically, this suggests that high-density montages may be redundant for identifying global abnormalities --- a finding with immediate implications for emergency or ICU settings where rapid assessment is required and full electrode setup is often impractical. For ECG, 41--75\% of leads are sufficient; dedicated single-lead validation in clinical settings remains an important future direction.

\paragraph{Limitations.}
The current evaluation covers EEG and ECG, modalities that are relatively proximal to PSG. Transfer to more distant modalities (e.g., EMG, respiration as a standalone task, BCI paradigms with high-density recording) and to non-clinical settings (wearables, resource-limited environments beyond BrainCapture) remains to be studied. The mixed validation protocols across datasets --- fixed splits for TUAB/TUEV, cross-validation elsewhere, external-transfer for ECG --- are motivated by established practice in each dataset's literature, but make direct cross-task statistical comparison less straightforward. Finally, comparisons for TUSL are necessarily limited because REVE, LaBraM, CBraMod, and BIOT do not report TUSL results in their publications and their checkpoints are not configured for this dataset.

\section{Conclusion}

This paper investigated whether sleep biosignals can serve as an effective pretraining distribution for adjacent biosignal domains recorded during wakefulness. We evaluated transfer performance across eight downstream tasks spanning five EEG and three ECG datasets. Our results demonstrate that sleep pretraining consistently improves performance over training from scratch and achieves state-of-the-art results on multiple EEG benchmarks. These findings validate that sleep's physiological richness --- long continuous contexts, diverse pathological waveforms, and cross-modal alignment --- enables learning of generalizable biosignal representations despite the low-density montages typical of polysomnography. Critically, our model achieves these results without any domain overlap with the wake EEG test sets, in contrast to competing models pretrained directly on TUH EEG data.

Beyond task performance, sleep-pretrained models maintain 95\% performance with 16--75\% of available channels and leads, demonstrating robustness to sensor availability without channel-specific design.

Our results support sleep as a promising pretraining distribution for biosignal foundation models, with particularly strong evidence in EEG and encouraging, but more limited, evidence in ECG external-transfer settings. Future work should investigate scaling sleep-pretrained models to larger parameter counts, combining sleep pretraining data with in-domain EEG corpora (e.g., REVE's 60,000+ hour corpus), and extending transfer evaluation to BCI paradigms and wearable biosignal settings.

\section*{Code Availability}
Code and trained models are available at \url{https://anonymous.4open.science/r/sleepfm4biosignals-3A50}.

\section*{Impact Statement}
This paper investigates whether sleep polysomnography can serve as an effective pretraining distribution for biosignal foundation models. Our goal is to inspire broader adoption of large-scale sleep datasets for pretraining, potentially improving clinical diagnostic tools for conditions such as epilepsy and cardiac arrhythmias while reducing the need for extensive labeled datasets in under-resourced healthcare settings.
As with all medical AI applications, downstream systems must be rigorously validated in clinical settings to ensure safety, fairness, and robustness across diverse patient populations before deployment. We encourage practitioners to evaluate models for potential biases, as sleep study cohorts may not fully represent the diversity of patients diagnosed during wakefulness.

\section*{Acknowledgements}

\bibliography{example_paper}
\bibliographystyle{icml2026}

\newpage
\appendix

\section{Appendix}

\subsection*{Dataset Demographics and Task Descriptions}

Demographics and task descriptions for the five EEG downstream dataset in Table \ref{tab:dataset_demographics}.

\begin{table*}[t!]
\centering
\caption{Demographics and task descriptions for the five EEG downstream datasets. TUEV and TUSL are subsets of the parent TUH EEG corpus; subject-level demographics are approximated from the full corpus (n = 10,874, $\sim$51\% female, mean age 51.6 $\pm$ 55.9 years). SPSW = spike and sharp wave; GPED = generalised periodic epileptiform discharges; PLED = periodic lateralised epileptiform discharges; EYEM = eye movement; ARTF = artifact; BCKG = background.}
\label{tab:dataset_demographics}
\small
\setlength{\tabcolsep}{4pt}
\renewcommand{\arraystretch}{1.15}
\begin{tabular}{p{1.8cm}p{1.8cm}p{0.9cm}p{1.5cm}p{1.0cm}p{5.5cm}}
\toprule
\textbf{Dataset} & \textbf{Task} & \textbf{N subj.} & \textbf{Age (mean $\pm$ SD)} & \textbf{\% F} & \textbf{Task description} \\
\midrule
TUAB \cite{lopez2015automated,obeid2016temple} &
Abnormality detection (2 classes) &
2,563 &
48.0 $\pm$ 17.9 yr (range: 7d--96 yr) &
30\% &
Binary classification of routine EEG ($\sim$20 min) as clinically normal or abnormal based on neurologist report labels. Pre-partitioned train/eval by subject (1,488 abnormal + 1,529 normal train; 126 abnormal + 150 normal eval). \\

TUEV \cite{harati2015improved,obeid2016temple} &
Event classification (6 classes) &
$\sim$518 files &
$\sim$51.6 $\pm$ 55.9 yr$^*$ &
$\sim$51\%$^*$ &
6-class classification of 5-second EEG segments into SPSW, GPED, PLED, EYEM, ARTF, or BCKG. Labels assigned by expert neurologists. Highly imbalanced. \\

TUSL \cite{shah2018temple,obeid2016temple} &
Slowing detection (4 classes) &
38 &
$\sim$51.6 $\pm$ 55.9 yr$^*$ &
$\sim$51\%$^*$ &
4-class classification for slowing events (Normal, Complex background, Slowing, Seizure). Relevant for encephalopathy and metabolic disorder monitoring. Highly imbalanced ($\sim$83\% normal). \\

CHB-MIT \cite{PhysioNet-chbmit-1.0.0,shoeb2009application} &
Seizure detection (2 classes) &
22 &
9.9 $\pm$ 5.0 yr (range: 1.5--22 yr) &
78.3\% &
Binary seizure detection in continuous scalp EEG from pediatric patients with intractable epilepsy. 198 annotated seizures; $<$1\% seizure prevalence (severe class imbalance). \\

BrainCapture \cite{armand2024low} &
Abnormality detection (2 classes) &
2,340 &
22.4 $\pm$ 18.9 yr (range: 1--60+ yr) &
41\% &
Binary normal/abnormal classification using a portable 27-channel EEG device in resource-limited settings in Kenya. Represents a non-Western, low-resource clinical population. \\
\bottomrule
\end{tabular}
\end{table*}

\subsection*{Extended ECG Results}

Extended tables for fine-tuned ECG models in Tables~\ref{tab:supp_ptbxl_results}, \ref{tab:supp_georgia_results}, and \ref{tab:supp_cpsc_results}.

\begin{table*}[t]
\centering
\caption{External-validation ECG transfer results after fine-tuning on Code-15: PTB-XL \smallheart\ dataset. We evaluate \model\ without pretraining (w/o PT) and with pretraining (w. PT).}
\label{tab:supp_ptbxl_results}
\begin{tabular}{lcccc} 
\hline
\textbf{Model} & \textbf{Model Size} & \textbf{Sensitivity} & \textbf{Specificity} & \textbf{F1} \\
\hline
CNN~\cite{narotamo2024deep} & 1.61M & $0.797\pm0.187$ & $0.994\pm0.003$ & $0.801\pm0.108$ \\
Resnet1d\_wang \cite{7966039} & 0.44M & $0.758\pm0.340$ & $0.992\pm0.004$ & $0.723\pm0.233$ \\
Xresnet1d101 \cite{8954382} & 28.3M & $0.844\pm0.204$ & $0.993\pm0.003$ & $0.803\pm0.107$ \\
Inception Time \cite{ismail_fawaz_inceptiontime_2020} & 0.45M & $0.842\pm0.184$ & $0.990\pm0.005$ & $0.771\pm0.089$ \\
Lstm \cite{narotamo2024deep} & 0.54M & $0.399\pm0.480$ & $0.991\pm0.012$ & $0.324\pm0.374$ \\
Lstm\_bidir \cite{narotamo2024deep} & 0.54M & $0.399\pm0.480$ & $0.991\pm0.012$ & $0.324\pm0.374$ \\
Wavelet+NN \cite{9190034} & 0.14M & $0.779\pm0.246$ & $0.994\pm0.003$ & $0.771\pm0.146$ \\
1DResNet \cite{ribeiro2020automatic} & 6.93M & $0.878\pm0.164$ & $0.988\pm0.007$ & $0.775\pm0.100$ \\
\hline
\model\ (w/o PT) & 3.58M & $0.780\pm0.097$ & $0.992\pm0.003$ & $0.782\pm0.066$ \\
\model\ (w. PT) & 3.58M & $0.792\pm0.090$ & $0.994\pm0.003$ & $0.807\pm0.071$ \\
\hline
\end{tabular}
\end{table*}

\begin{table*}[t]
\centering
\caption{External-validation ECG transfer results after fine-tuning on Code-15: Georgia \smallheart\ dataset. We evaluate \model\ without pretraining (w/o PT) and with pretraining (w. PT).}
\label{tab:supp_georgia_results}
\begin{tabular}{lcccc}
\hline
\textbf{Model} & \textbf{Model Size} & \textbf{Sensitivity} & \textbf{Specificity} & \textbf{F1} \\
\hline
CNN \cite{narotamo2024deep}& 1.61M & $0.636\pm0.113$ & $0.989\pm0.011$ & $0.689\pm0.081$ \\
Resnet1d\_wang \cite{7966039} & 0.44M & $0.724\pm0.161$ & $0.989\pm0.004$ & $0.728\pm0.070$ \\
Xresnet1d101 \cite{8954382} & 28.30M & $0.776\pm0.103$ & $0.988\pm0.004$ & $0.763\pm0.028$ \\
Inception Time \cite{ismail_fawaz_inceptiontime_2020} & 0.45M & $0.800\pm0.093$ & $0.983\pm0.004$ & $0.736\pm0.056$ \\
Lstm \cite{narotamo2024deep} & 0.54M & $0.364\pm0.431$ & $0.990\pm0.012$ & $0.313\pm0.362$ \\
Lstm\_bidir \cite{narotamo2024deep} & 0.54M & $0.364\pm0.431$ & $0.990\pm0.012$ & $0.313\pm0.362$ \\
Wavelet+NN \cite{9190034} & 0.14M & $0.613\pm0.201$ & $0.991\pm0.006$ & $0.663\pm0.144$ \\
1DResNet \cite{ribeiro2020automatic} & 6.93M & $0.803\pm0.103$ & $0.983\pm0.002$ & $0.737\pm0.041$ \\
\hline
\model\ (w/o PT) & 3.02M & $0.729\pm0.068$ & $0.979\pm0.011$ & $0.691\pm0.045$ \\
\model\ (w. PT) & 3.02M & $0.732\pm0.064$ & $0.983\pm0.010$ & $0.713\pm0.023$ \\
\hline
\end{tabular}
\end{table*}

\begin{table*}[t]
\centering
\caption{External-validation ECG transfer results after fine-tuning on Code-15: CPSC \smallheart\ dataset. We evaluate \model\ without pretraining (w/o PT) and with pretraining (w. PT).}
\label{tab:supp_cpsc_results}
\begin{tabular}{lcccc}
\hline
\textbf{Model} & \textbf{Model Size} & \textbf{Sensitivity} & \textbf{Specificity} & \textbf{F1} \\
\hline
CNN \cite{narotamo2024deep} & 1.61M & $0.616\pm0.188$ & $0.991\pm0.008$ & $0.711\pm0.127$ \\
Resnet1d\_wang \cite{7966039} & 0.44M & $0.670\pm0.166$ & $0.991\pm0.005$ & $0.748\pm0.113$ \\
Xresnet1d101 \cite{8954382} & 28.30M & $0.685\pm0.155$ & $0.990\pm0.005$ & $0.758\pm0.101$ \\
Inception Time \cite{ismail_fawaz_inceptiontime_2020} & 0.45M & $0.707\pm0.099$ & $0.989\pm0.005$ & $0.770\pm0.070$ \\
Lstm \cite{narotamo2024deep} & 0.54M & $0.290\pm0.353$ & $0.994\pm0.007$ & $0.314\pm0.363$ \\
Lstm\_bidir \cite{narotamo2024deep} & 0.54M & $0.290\pm0.353$ & $0.994\pm0.007$ & $0.314\pm0.363$ \\
Wavelet+NN \cite{9190034} & 0.14M & $0.630\pm0.205$ & $0.992\pm0.007$ & $0.719\pm0.144$ \\
1DResNet \cite{ribeiro2020automatic} & 6.93M & $0.738\pm0.134$ & $0.989\pm0.004$ & $0.784\pm0.086$ \\
\hline
\model\ (w/o PT) & 3.02M & $0.724\pm0.092$ & $0.980\pm0.010$ & $0.759\pm0.071$ \\
\model\ (w. PT) & 3.02M & $0.760\pm0.096$ & $0.980\pm0.096$ & $0.770\pm0.068$ \\
\hline
\end{tabular}
\end{table*}

\begin{table*}[t]
\centering
\caption{Recent ECG literature relevant to our setup. Direct comparison is limited because our ECG study fine-tunes on Code-15 and evaluates on external datasets with aligned arrhythmia labels.}
\label{tab:ecg_lit}
\small
\setlength{\tabcolsep}{4pt}
\renewcommand{\arraystretch}{1.15}
\begin{tabular}{p{3.1cm}p{1.4cm}p{3.6cm}p{1.8cm}p{4.2cm}p{1.7cm}}
\toprule
\textbf{Work} & \textbf{Type} & \textbf{Protocol} & \textbf{Metric} & \textbf{Reported value(s)} & \textbf{Directly comparable?} \\
\midrule

Strodthoff et al.\ /\newline PTB-XL benchmark \cite{9190034,ptbxl_benchmark_repo}
& Supervised
& Standard PTB-XL rhythm benchmark
& AUC
& xresnet1d101: 0.957;\newline inception1d: 0.953;\newline LSTM: 0.953;\newline resnet1d\_wang: 0.946
& No \\

OpenECG \cite{openecg2025}
& ECG-FM benchmark
& Multi-dataset public ECG benchmark with unified 5-fold evaluation
& F1 /\newline AUROC
& PTB-XL: BYOL 47.7 / 91.1,\newline MAE 48.1 / 90.9;\newline CPSC: MAE 74.5 / 93.2;\newline Georgia: BYOL 26.2 / 68.5,\newline MAE 25.3 / 67.9
& No \\

ECG-FM \cite{mckeen2025ecg}
& ECG-FM
& Large-scale pretraining with downstream institutional tasks
& AUROC
& AF: 0.996;\newline LVEF: 0.929
& No \\

KED \cite{ked2024}
& ECG-FM
& Zero-shot / few-shot diagnosis across regions
& AUC
& Conduction-block avg.: 0.921;\newline premature contractions avg.: 0.944
& No \\
\bottomrule
\end{tabular}
\end{table*}
\onecolumn



\end{document}